\def\year{2020}\relax
\documentclass[letterpaper]{article} 
\usepackage{aaai20}  
\usepackage{times}  
\usepackage{helvet} 
\usepackage{courier}  
\usepackage[hyphens]{url}  
\usepackage{graphicx} 
\usepackage{graphicx}  
\newcommand{\citet}[1]{\citeauthor{#1} \shortcite{#1}}
\newcommand{\citep}{\cite}

\usepackage{amssymb}
\usepackage{booktabs}
\usepackage{subcaption}
\newcommand{\g}{\, | \,}
\newcommand{\abr}[1]{\textsc{#1}}

\usepackage{bm}
\newcommand{\vect}[1]{\bm{\mathbf{#1}}}

\newcommand{\name}[0]{\textsc{caco}}

\newcommand{\pp}[0]{\textsuperscript{p}}

\usepackage{mathtools}
\DeclarePairedDelimiterX{\infdivx}[2]{(}{)}{%
  #1\;\delimsize\|\;#2%
}
\newcommand{\kldiv}{\mathrm{KL}\infdivx}
\DeclarePairedDelimiter{\norm}{\lVert}{\rVert}

\usepackage{array}
\newcolumntype{R}[0]{>{\raggedleft\let\newline\\\arraybackslash\hspace{0pt}}p{1cm}}

\newcommand{\flag}[1]{\includegraphics{#1.pdf}}

\title{Exploiting Cross-Lingual Subword Similarities in Low-Resource Document
Classification}

\author{Mozhi Zhang \\
  \textsc{cs} and \textsc{umiacs} \\
  University of Maryland \\
  College Park, MD, USA \\
  mozhi@cs.umd.edu
  \And
  Yoshinari Fujinuma \\
  Computer Science \\
  University of Colorado \\
  Boulder, CO, USA \\
  fujinumay@gmail.com
  \And
  Jordan Boyd-Graber\thanks{Now at Google Research Z\"urich} \\
  \textsc{cs}, iSchool, \textsc{lsc}, and \textsc{umiacs}\\
  University of Maryland \\
  College Park, MD, USA \\
  jbg@umiacs.umd.edu \\}

\begin{document}
\maketitle
\begin{abstract}

  Text classification must sometimes be applied in a low-resource language with
  no labeled training data.
  However, training data may be available in a \emph{related} language.
  We investigate whether character-level knowledge transfer from a related
  language helps text classification.
  We present a cross-lingual document classification framework (\name{}) that
  exploits cross-lingual subword similarity by jointly training a
  character-based embedder and a word-based classifier.
  The embedder derives vector representations for input words from their
  written forms, and the classifier makes predictions based on the word
  vectors.
  We use a joint character representation for both the source language and the
  target language, which allows the embedder to generalize knowledge about
  source language words to target language words with similar forms.
  We propose a multi-task objective that can further improve the model if
  additional cross-lingual or monolingual resources are available.
  Experiments confirm that character-level knowledge transfer is more
  data-efficient than word-level transfer between related languages.

\end{abstract}

\begin{figure*}
  \centering
  \includegraphics[width=.9\textwidth]{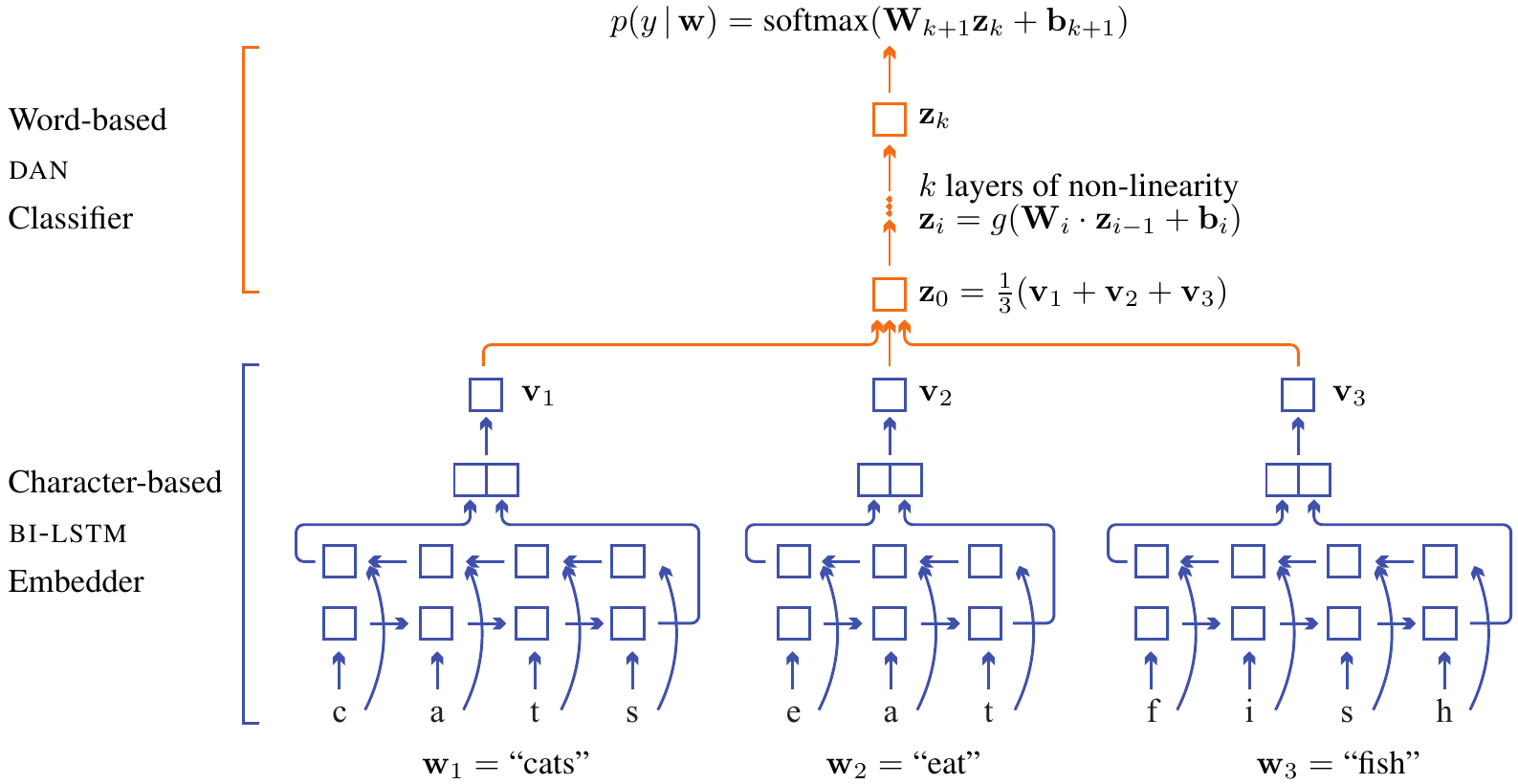}
  \caption{\label{fig:model} Computation graph of \name{} on an example
  sentence (``cats eat fish'').
  \emph{Bottom:} Each input word $\vect{w}_i$ is mapped to a vector $\vect{v}_i$
  by passing its characters through a \abr{bi-lstm} embedder.
  \emph{Top:} Word vectors $\{\vect{v_i}\}$ are then passed through a \abr{dan}
  classifier to predict the label $y$.
  Specifically, \abr{dan} transforms the average of the word vectors
  $\vect{z}_0$ with $k$ layers of non-linearity and a final softmax layer.}
\end{figure*}

\section{Introduction: Classifiers across Languages}

Modern machine learning methods in natural language processing can learn highly
accurate, context-based classifiers~\citep{devlin-19}.
Despite this revolution for high-resource languages such as English, some
languages are left behind because of the dearth of text data generally and
specifically labeled data.
Often, the need for a text classifier in a low-resource language is acute, as
text classifiers can provide situational awareness in emergent
incidents~\citep{strassel-16}.
Cross-lingual document
classification~\citep[\abr{cldc}]{klementiev-12} attacks this problem
by using annotated dataset from a \emph{source} language to build
classifiers for a \emph{target} language.

\abr{cldc} works when it can find a shared representation for
documents from both languages: train a classifier on source language
documents and apply it on target language documents.
Previous work uses a bilingual lexicon~\citep{shi-10,andrade-15}, machine
translation~\citep[\abr{mt}]{banea-08-fixed,wan-09-fixed,zhou-16}, topic
models~\citep{mimno-09-fixed,yuan-18}, cross-lingual word
embeddings~\citep[\abr{clwe}]{klementiev-12}, or multilingual contextualized
embeddings~\citep{wu-19} to extract cross-lingual features.
But these methods may be impossible in low-resource languages, as they require
some combination of large parallel or comparable text, high-coverage
dictionaries, and monolingual corpora from a shared domain.

However, as anyone who has puzzled out a Portuguese menu from their
high school Spanish knows, the task is not hopeless, as languages do
not exist in isolation.
Shared linguistic roots, geographic proximity, and history bind
languages together; cognates abound, words sound the same, and there
are often shared morphological patterns.
These similarities are often not found at word-level but at character-level.
Therefore, we investigate character-level knowledge transfer for \abr{cldc} in
truly low-resource settings, where unlabeled or parallel data in the target
language is also limited or unavailable.

To study knowledge transfer at character level, we propose a \abr{cldc}
framework, {\bf C}lassification {\bf A}ided by {\bf C}onvergent {\bf
O}rthography (\name{}) that capitalizes on character-level similarities between
related language pairs.
Previous \abr{cldc} methods treat words as atomic symbols and do not transfer
character-level patterns across languages; \name{} instead uses a bi-level
model with two components: a character-based \emph{embedder} and a word-based
\emph{classifier}.

The embedder exploits shared patterns in related languages to
create word representations from character sequences.
The classifier then uses the \emph{shared} representation across
languages to label the document.  
The embedder learns morpho-semantic regularities, while the
classifier connects lexical semantics to labels.

To allow cross-lingual transfer, we use a \emph{single} model with shared
character embeddings for both languages.
We jointly train the embedder and the classifier on annotated source
language documents.
The embedder transfers knowledge about source language words to target language
words with similar orthographic features.

While the model can be fairly accurate without any target language data, it can
also benefit from \emph{a small amount} of additional information when
available.  If we have a dictionary, pre-trained monolingual word embeddings,
or parallel text, we can fine-tune the model with multi-task learning.
We encourage the embedder to produce similar word embeddings for translation
pairs from a dictionary, which captures patterns between cognates.
We also teach the embedder to \abr{mimick} pre-trained word embeddings
in the source language~\citep{pinter-17}, which exposes the model to
more word types.
When we have a good reference model in another high-resource language, we can
train our model to make similar predictions as the reference model on
parallel text~\citep{xu-17}.

We verify the effectiveness of character-level knowledge transfer on two
\abr{cldc} benchmarks.
When we have enough data to learn high-quality \abr{clwe}, training classifiers
with \abr{clwe} as input features is a strong \abr{cldc} baseline.
\name{} can match the accuracy of \abr{clwe}-based models \emph{without}
using any target language data, and fine-tuning the embedder with a
small amount of additional resources improves \name{}'s accuracy.
Finally, \name{} is also useful when we have enough resources to train
good \abr{clwe}---using \abr{clwe} as extra features, \name{} outperforms the
baseline \abr{clwe}-based models by a large margin.

\section{\name{}: Classification Aided by Convergent Orthography}
\label{sec:model}

This section introduces our method, \name{}, which trains a multilingual
document classifier using labeled datasets in a source language $\mathcal{S}$
and applies the classifier to a low-resource target language $\mathcal{T}$.
We focus on the setting where $\mathcal{S}$ and $\mathcal{T}$ are related and
have similar orthographic features.

\subsection{Model Architecture}

Let $\vect{x}$ be an input document with a sequence of words $\vect{x} =
\langle \vect{w}_1, \vect{w}_2, \cdots, \vect{w}_n \rangle$, where each word
$\vect{w}_i$ is a sequence of character.
Our model maps the document~$\vect{x}$ to a distribution over possible
labels~$y$ in two steps (Figure~\ref{fig:model}).
First, we generate a word embedding~$\vect{v}_i$ for each input
word~$\vect{w}_i$ using a character-based embedder~$e$:
\begin{equation}
  \vect{v}_i = e(\vect{w}_i).
\end{equation}
We then feed the word embeddings to a word-based classifier~$f$ to compute the
distribution over labels $y$:
\begin{equation}
  p(y \g \vect{w}) = f\left( \langle \vect{v}_1, \vect{v}_2, \cdots, \vect{v}_n \rangle \right ).
\end{equation}
We can use any sequence model for the embedder~$e$ and the classifier~$f$.  For
our experiments, we use a bidirectional \abr{lstm}
~\citep[\abr{bi-lstm}]{graves-05} embedder and a deep averaging network
~\citep[\abr{dan}]{iyyer-15-fixed} classifier.

\paragraph{\abr{bi-lstm} Embedder.}
\abr{bi-lstm} is a powerful sequence model that captures complex non-local
dependencies.  Character-based \abr{bi-lstm} embedders are used in
many natural language processing
tasks~\citep{ling-15a,ballesteros-15,lample-16}.
To embed a word $\vect{w}$, we pass the character sequence~$\vect{w}$ to a
left-to-right \abr{lstm} and the reversed character sequence~$\vect{w'}$ to a
right-to-left \abr{lstm}.  We concatenate the final hidden states of the two
\abr{lstm} and apply a linear transformation:
\begin{equation}
  e(\vect{w}) = \vect{W}_e \cdot [\overrightarrow{\text{\abr{lstm}}}(\vect{w});
  \overleftarrow{\text{\abr{lstm}}}(\vect{w'})] + \vect{b}_e,
\end{equation}
where the functions $\overrightarrow{\text{\abr{lstm}}}$ and
$\overleftarrow{\abr{lstm}}$ compute the final hidden states of the two
\abr{lstm}s.

\paragraph{\abr{dan} Classifier.}
A \abr{dan} is an unordered model that passes the arithmetic mean of the input
word embeddings through a multilayer perceptron and feeds the final layer's
representation to a softmax layer.
\abr{dan} ignores cross-lingual variations in word order (i.e., syntax) and
thus generalizes well in \abr{cldc}.
Despite its simplicity, \abr{dan} has near state-of-the-art accuracies on both
monolingual~\citep{iyyer-15-fixed} and cross-lingual document
classification~\citep{chen-18}.

Let  $\vect{v}_1,\vect{v}_2,\cdots,\vect{v}_n$ be the word embeddings
generated by the character-based embedder.
\abr{dan} uses the average of the word embeddings as the document
representation $\vect{z}_0$:
\begin{equation}
  \vect{z}_0 = \frac{1}{n}\sum_{i=1}^n \vect{v}_i,
\end{equation}
and $\vect{z}_0$ is passed through $k$ layers of non-linearity:
\begin{equation}
  \vect{z}_i = g(\vect{W}_i \cdot \vect{z}_{i-1} + \vect{b}_i),
\end{equation}
where $i$ ranges from 1 to $k$, and $g$ is a non-linear activation
function.
The final representation~$\vect{z}_{k}$ is passed to a softmax layer to obtain
a distribution over the label $y$,
\begin{equation}
    p(y \g \vect{x}) = \text{softmax}(\vect{W}_{k+1} \vect{z}_k + \vect{b}_{k+1}).
\end{equation}

We use the same classifier parameters $\vect{W}_i$ across languages.
In other words, the \abr{dan} classifier is language-independent.
This is possible because the embedder generates consistent word representations
across related languages, which we discuss in the next section.

\subsection{Character-Level Cross-Lingual Transfer}
\label{ssec:char}

To transfer character-level information across languages, the embedder uses
the same character embeddings for both languages.
The character-level \abr{bi-lstm} vocabulary is the union of the alphabets for
the two languages, and the embedder does not differentiate identical characters
from different languages.
For example, a Spanish ``a'' has the same character embedding as a French
``a''.
Consequently, the embedder maps words with similar forms from both languages to
similar vectors.

If the source language and the target language are orthographically similar, the
embedder can generalize knowledge learned about source language words to target
language words through shared orthographic features.  As an example, if the
model learns that the Spanish word ``religioso'' (religious) is predictive of
label~$y$, the model automatically infers that ``religioso'' in Italian is also
predictive of $y$, even though the model never sees any Italian text.

In our experiments, we focus on related language pairs that share the same
script.
For related languages with different scripts, we can apply \name{} to the
output of a transliteration tool or a grapheme-to-phoneme
transducer~\citep{mortensen-18}.
We leave this to future work.

\subsection{Training Objective}\label{ssec:objective}

Our main objective is supervised document classification.  We jointly
train the classifier and the embedder to minimize average negative
log-likelihood on labeled source language documents $S$:
\begin{equation}
  L_s(\vect{\vect{\theta}}) = -\frac{1}{|S|}\sum_{\langle \vect{x}, y \rangle}
  \log p(y \g \vect{x}),
\end{equation}
where $\vect{\theta}$ is a vector representing all model parameters, and $S$ is
a set of source language examples with words~$\vect{x}$ and label~$y$.

Sometimes we have additional resources for the source or target language.  We
use them to improve \name{} with multi-task learning~\citep{collobert-11b}
via three auxiliary tasks.  

\paragraph{Word Translation (\abr{dict}).}
There are many patterns when translating cognate words between related
languages.
For example, Italian ``e'' often becomes ``ie'' in Spanish.
``Tempo'' (time) in Italian becomes ``tiempo'' in Spanish, and ``concerto''
(concert) in Italian becomes ``concierto'' in Spanish.
The embedder can learn these word translation patterns from a bilingual
dictionary.

Let $D$ be a bilingual dictionary with a set of word pairs $\langle \vect{w}_s,
\vect{w}_t \rangle$, where $\vect{w}_s$ and $\vect{w}_t$ are translations of
each other. 
We add a term to our objective to minimize average squared Euclidean distances
between the embeddings of translation pairs~\citep{mikolov-13b}:
\begin{equation}
  L_d(\vect{\theta}) = \frac{1}{|D|}\sum_{\langle \vect{w}_s, \vect{w}_t \rangle}
  \norm{e(\vect{w}_s) - e(\vect{w}_t)}_2^2.
\end{equation}

\paragraph{Mimicking Word Embeddings (\abr{mim}).}
Monolingual text classifiers often benefit from initializing embeddings with
word vectors pre-trained on large unlabeled corpus~\citep{collobert-11b}.  This
semi-supervised learning strategy helps the model generalize to word types
outside labeled training data.  Similarly, our embedder can
\abr{mimick}~\citep{pinter-17} an existing \emph{source language}
word embeddings to generalize better.

Suppose we have a pre-trained source language word embedding matrix $\vect{E}$
with $V$ rows.  The $i$-th row $\vect{x}_i$ is a vector for the $i$-th word
type $\vect{w}_i$.
We add an objective to minimize the average squared Euclidean distances between
the output of the embedder and $\vect{E}$:
\begin{equation}
  L_e(\vect{\theta}) = \frac{1}{V} \sum_{i=1}^V \norm{e(\vect{w}_i) - \vect{E}_i}_2^2.
\end{equation}

\begin{table*}
  \centering
  \begin{tabular}{lccccccc}
    \toprule
      & \multicolumn{4}{c}{\name{}} & \multicolumn{2}{c}{Baseline}\\
    \cmidrule(lr){2-5} \cmidrule(lr){6-7}
    & \abr{src} & \abr{dict} & \abr{mim} & \abr{all} & \abr{clwe} & \abr{sup} & \abr{com}\\
    \midrule
    Source labeled data & \checkmark & \checkmark & \checkmark & \checkmark & \checkmark & \checkmark & \checkmark\\
    Pre-trained source embedding & & & \checkmark & \checkmark \\
    Small dictionary & & \checkmark & & \checkmark \\
    Pre-trained \abr{clwe} & & & & & \checkmark & & \checkmark\\
    Target labeled data & & & & & & \checkmark\\
    \midrule
    \abr{rcv2} average accuracy & 50.0 & 55.7 & 51.5 & 54.7 & 51.6 & 51.9 & \textbf{64.5} \\
    \bottomrule
  \end{tabular}
  \caption{Comparison of models used in our experiments (introduced in
  Section~\ref{ssec:model}).
  For each model, we list its required resources and average accuracy on
  \abr{rcv2} over eight related language pairs (accuracy for each pair in
  Table~\ref{tab:rcv2}).
  We compare \name{} variants with two high-resource models: a \abr{clwe}-based
  model (\abr{clwe}) and a lightly supervised target language model
  (\abr{sup}).
  Both baselines require more target language resources than \name{} variants,
  and yet they have lower average accuracy than some \name{} variants, which
  confirms that character-level knowledge transfer is highly efficient.
  We also experiment with a model that combines \abr{clwe} with \name{} (\abr{com}).
  This combined model has the highest average accuracy, indicating that
  \abr{clwe} and \name{} are complementary when both options are available.
  } 
  \label{tab:model}
\end{table*}

\paragraph{Knowledge Distillation.}
Sometimes we have a reliable reference classifier in another high-resource
language $\mathcal{H}$ (e.g., English).  If we have parallel text
between $\mathcal{S}$ and $\mathcal{H}$, we can use knowledge
distillation~\citep{xu-17} to supply additional training signal.  Let $P$
be a set of parallel documents $\langle \vect{x}_s, \vect{x}_h \rangle$, where
$\vect{x}_s$ is from source language $\mathcal{S}$, and $\vect{x}_h$ is the
translation of $\vect{x}_s$ in $\mathcal{H}$.  We add another objective term to
minimize the average Kullback-Leibler divergence between the predictions of our
model and the reference model:
\begin{equation}
  L_p(\vect{\theta}) = \frac{1}{|P|} \sum_{\langle \vect{x}_s, \vect{x}_h \rangle \in P}
  \kldiv{p_h (y \g \vect{x}_h)}{p(y \g \vect{x}_s)},
\end{equation}
where $p_h$ is the output of the reference classifier (in language
$\mathcal{H})$, and $p$ is the output of \name{}.  In \S~\ref{sec:experiments},
we mark models that use knowledge distillation with a superscript~``\abr{p}''.

We train on the four tasks jointly.  Our final objective is:
\begin{equation}
  L(\vect{\theta}) = L_s(\vect{\theta}) + \lambda_d L_d(\vect{\theta}) + \lambda_e L_e(\vect{\theta}) + \lambda_p L_p(\vect{\theta}),\label{eq:full_obj}
\end{equation}
where the hyperparameters $\lambda_d$, $\lambda_e$, and $\lambda_p$ trade off
between the four tasks.

\section{Experiments}\label{sec:experiments}

When the source language and the target language are related, we expect
character-level knowledge transfer to be more data-efficient than word-level
knowledge transfer because character-level transfer allows generalization
across words with similar forms. 
We test this by comparing \name{} models trained in low-resource settings and
with \abr{clwe}-based models trained in high-resource settings on two
\abr{cldc} datasets.
We also compare \name{} with a supervised monolingual model.
On both datasets, \name{} models have similar average accuracy as the
baselines \emph{while requiring much less target language data}.
Finally, we train models that combine \name{} with \abr{clwe}, which have
significantly higher accuracy than models with only \abr{clwe} as features.
These results confirms that character-level similarities between related
languages effectively transfer knowledge for \abr{cldc}.

\subsection{Classification Dataset} 

Our first dataset is Reuters multilingual corpus (\abr{rcv2}), a collection of
news stories labeled with four topics~\citep{lewis-04}:Corporate/Industrial
(\abr{ccat}), Economics (\abr{ecat}), Government/Social (\abr{gcat}), and
Markets (\abr{mcat}).
Following \citet{klementiev-12}, we remove documents with multiple topic
labels.  For each language, we sample 1,500 training documents and 200 test
documents with balanced labels.  We conduct \abr{cldc} experiments between two
North Germanic languages, Danish~(\abr{da}) and Swedish~(\abr{sv}), and three
Romance languages, French~(\abr{fr}), Italian~(\abr{it}), and
Spanish~(\abr{es}).

To test \name{} on truly low-resource languages, we build a second \abr{cldc}
dataset with famine-related documents sampled from Tigrinya~(\abr{ti}) and
Amharic~(\abr{am}) \abr{lorelei} language packs~\citep{strassel-16}.
We train binary classifiers to detect whether the document describes widespread
crime or not.  For Tigrinya documents, the labels are extracted from the
situation frame annotation in the language pack.  We mark all documents with a
``widespread crime/violence'' situation frame as positive.  The Amharic
language pack does not have annotations, so we label Amharic sentences
based on English reference translations included from the language pack.
Our dataset contains 394 Tigrinya and 370 Amharic documents with balanced
labels.

\begin{table*}
  \centering
  \begin{tabular}{llccccccc}
    \toprule
    & & \multicolumn{4}{c}{\name{}}\\
    \cmidrule(lr){3-6}
    source & target & \abr{src} & \abr{dict} & \abr{mim} & \abr{all} & \abr{clwe} & \abr{sup} & \abr{com}\\
    \midrule
    \flag{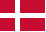}~\abr{da} & \flag{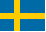}~\abr{sv} & 56.0 & 62.8 & 60.4 & 62.9 & 69.3 & 59.7 & {\bf 69.7} \\
    \flag{sv}~\abr{sv} & \flag{da}~\abr{da} & 56.7 & 60.2 & 58.4 & 62.2 & 51.4 & 40.8 & {\bf 67.5} \\
    \flag{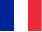}~\abr{fr} & \flag{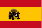}~\abr{es} & 49.6 & 59.3 & 48.3 & 57.4 & 63.9 & 56.6 & {\bf 70.8} \\
    \flag{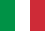}~\abr{it} & \flag{es}~\abr{es} & 50.2 & 54.6 & 51.4 & 54.7 & 43.4 & 56.6 & {\bf 63.5} \\
    \flag{es}~\abr{es} & \flag{fr}~\abr{fr} & 48.5 & 49.7 & 49.2 & 48.8 & {\bf 63.1} & 48.9 & 61.3 \\
    \flag{it}~\abr{it} & \flag{fr}~\abr{fr} & 45.9 & 52.1 & 46.6 & 48.2 & 26.7 & 48.9 & {\bf 62.8} \\
    \flag{fr}~\abr{fr} & \flag{it}~\abr{it} & 43.3 & 53.2 & 44.3 & 51.2 & 43.6 & 44.9 & {\bf 60.2} \\
    \flag{es}~\abr{es} & \flag{it}~\abr{it} & 49.7 & 53.5 & 53.4 & 52.5 & 51.3 & 44.9 & {\bf 59.7} \\
    \multicolumn{2}{r}{average} & 50.0 & 55.7 & 51.5 & 54.7 & 51.6 & 51.9 & {\bf 64.5} \\
    \bottomrule
  \end{tabular}
  \caption{\abr{cldc} experiments between eight related European language pairs
  on \abr{rcv2} topic identification.
  The average accuracy of \name{} models are competitive with word-based models
  that use \emph{more resources} such as target language corpora or labeled
  data (Table~\ref{tab:model}).
  The combined model (\abr{com}) has the highest
  average test accuracy.  We \textbf{boldface} the best result for each row.}
  \label{tab:rcv2} 
\end{table*}

\begin{table*}
  \centering
  \begin{tabular}{llcccccc}
    \toprule
    & & \multicolumn{4}{c}{\name{}}\\
    \cmidrule(lr){3-6}
    source & target & \abr{src} & \abr{mim} & \abr{src\pp{}} & \abr{mim\pp{}} & \abr{clwe} & \abr{com}\\
    \midrule
    \flag{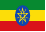}~\abr{am} & \flag{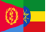}~\abr{ti} & 55.5 & 56.3 & 57.0 & 57.6  & 59.1 & {\bf 60.1} \\
    \flag{ti}~\abr{ti} & \flag{am}~\abr{am} & 56.8 & 55.1 & * & * & 58.1 & {\bf 59.5} \\
    \bottomrule
  \end{tabular}
  \caption{\abr{cldc} experiments between Amharic and Tigrinya on \abr{lorelei}
  disaster response dataset.  \name{} models are only slightly worse than
  \abr{clwe}-based models without using any target language data.
  For \abr{am}-\abr{ti}, knowledge distillation (\abr{src\pp{}} and
  \abr{mim\pp{}}) further improves \name{} models.
  We do not experiment with knowledge distillation on \abr{ti}
  because we cannot find enough unlabeled parallel text in the language pack.
  Combining \name{} with pre-trained \abr{clwe} gives the highest test accuracy.}
  \label{tab:lorelei} 
\end{table*}

\subsection{Models}
\label{ssec:model}

We compare \name{} trained under low-resource settings with word-based models
that use more resources.
Table~\ref{tab:model} summarizes our models.

\paragraph{\name{} Variants.}
We experiment with several variants of \name{} that uses different resources.
The \textbf{\abr{src}} model uses the least amount of resource.
It is only trained on labeled source language documents and do not use any
unlabeled data.
The \textbf{\abr{dict}} model requires a dictionary and is trained with the
word translation auxiliary task.
The \textbf{\abr{mim}} model requires a pre-trained source language embedding
and uses the mimick auxliliary task.
The \textbf{\abr{all}} model is the most expensive variant.  It is trained with
both the word translation and the mimick auxiliary tasks.  
In \abr{lorelei} experiments, we also use knowledge distillation to provide
more classification signals for some models.
We mark these models with a superscript~``\abr{p}''.

\paragraph{\abr{clwe}-Based Model.}
Our first word-based model is a \abr{dan} with pre-trained multiCCA \abr{clwe}
features~\citep{ammar-16}.
The \abr{clwe} are trained on large target language corpora with millions of
tokens and high-coverage dictionaries with hundreds of thousands of word types.
In contrast, we train \name{} models in a simulated low-resource setting with
few or no target language data.
Despite the resource gap, \name{} models have similar average test accuracy as
\abr{clwe}-based models, demonstrating the effectiveness of character-level
transfer learning.

\paragraph{Supervised Model.}
Next, we compare \name{} with a lightly-supervised monolingual model
(\textbf{\abr{sup}}), a word-based \abr{dan} trained on fifty labeled target
language documents.
We only apply this baseline to \abr{rcv2}, because the labeled document sets in
\abr{lorelei} are too small to split further.
The supervised model requires labeled target language documents, which often
do not exist in labeled documents.
Without using any target language supervision, \name{} models have similar
(and sometimes higher) test accuracies as \abr{sup}, showing that \name{}
effectively learns from a related language.

\paragraph{Combined Model.}
Finally, we experiment with a model that combines \name{} and
\abr{clwe} (\textbf{\abr{com}}) by feeding pre-trained \abr{clwe} as
additional features for the classifier of a \name{} model (\abr{src}
variant).
This model requires the same amount of resource as the \abr{clwe}-based model.
The combined model on average has much higher accuracy than
both \abr{caco} variants and \abr{clwe}-based model, showing that
character-level knowledge transfer is useful even when we have enough
unlabeled data to train high-quality \abr{clwe}.

\begin{table*}
  \tabcolsep=0.12cm
  \centering
  \begin{subtable}{.47\linewidth}
    \centering
    \begin{tabular}{llccccc}
      \toprule
      & & \multicolumn{4}{c}{\name{}}\\
      \cmidrule(lr){3-6}
      source & target & \abr{src} & \abr{dict} & \abr{mim} & \abr{all} & \abr{clwe} \\
      \midrule
      \flag{da}~\abr{da} & \flag{es}~\abr{es} & 32.5 & 34.8 & 30.6 & 38.2 & {\bf 65.7} \\
      \flag{da}~\abr{da} & \flag{fr}~\abr{fr} & 34.1 & 41.8 & 35.5 & 43.3 & {\bf 45.9} \\
      \flag{da}~\abr{da} & \flag{it}~\abr{it} & 36.8 & 43.7 & 37.2 & 41.5 & {\bf 47.4} \\
      \flag{sv}~\abr{sv} & \flag{es}~\abr{es} & 35.2 & 42.5 & 34.6 & 46.8 & {\bf 48.5} \\
      \flag{sv}~\abr{sv} & \flag{fr}~\abr{fr} & 27.4 & 29.9 & 29.1 & 28.3 & {\bf 49.0} \\
      \flag{sv}~\abr{sv} & \flag{it}~\abr{it} & 34.6 & 36.4 & 33.3 & 35.2 & {\bf 40.4} \\
      \multicolumn{2}{r}{average} & 33.4 & 38.2 & 33.4 & 37.2 & {\bf 49.5} \\
      \bottomrule
    \end{tabular}
    \caption{North Germanic to Romance}
  \end{subtable}
  \hspace{1em}
  \begin{subtable}{.47\linewidth}
    \centering
    \begin{tabular}{llccccc}
      \toprule
      & & \multicolumn{4}{c}{\name{}}\\
      \cmidrule(lr){3-6}
      source & target & \abr{src} & \abr{dict} & \abr{mim} & \abr{all} & \abr{clwe} \\
      \midrule
      \flag{es}~\abr{es} & \flag{da}~\abr{da} & 47.7 & 48.3 & 46.1 & 52.0 & {\bf 56.7} \\
      \flag{es}~\abr{es} & \flag{sv}~\abr{sv} & 50.6 & {\bf 53.7} & 48.5 & 51.4 & 52.4 \\
      \flag{fr}~\abr{fr} & \flag{da}~\abr{da} & 46.7 & 44.2 & 44.7 & {\bf 48.6} & 45.3 \\
      \flag{fr}~\abr{fr} & \flag{sv}~\abr{sv} & 52.9 & 53.2 & 53.6 & 52.8 & {\bf 57.2} \\
      \flag{it}~\abr{it} & \flag{da}~\abr{da} & 36.6 & 43.6 & 34.8 & 43.0 & {\bf 48.2} \\
      \flag{it}~\abr{it} & \flag{sv}~\abr{sv} & 37.8 & {\bf45.3} & 30.7 & 43.9 & 31.1 \\
      \multicolumn{2}{r}{average} & 45.4 & 48.1 & 43.1 & {\bf 48.6} & 48.5 \\
      \bottomrule
    \end{tabular}
    \caption{Romance to North Germanic}
  \end{subtable}
  \caption{\abr{cldc} experiments between languages from different families on
  \abr{rcv2}.  When transferring from a North Germanic language to a Romance
  language, \name{} models score much lower than \abr{clwe}-based models (left).
  Surprisingly, \name{} models are on par with \abr{clwe}-based when
  transferring from a Romance language to a North Germanic language (right).
  We \textbf{boldface} the best result for each row.}
  \label{tab:unrelated} 
\end{table*}

\subsection{Auxiliary Task Data}

Some of the \name{} models (\abr{dict} and \abr{all}) use a dictionary to learn
word translation patterns.
We train them on the same training dictionary used for pre-training the
\abr{clwe}.
To simulate the low-resource setting, we sample \textbf{only 100 translation
pairs} from the original dictionary for \name{}.
Pilot experiments confirm that a larger dictionary can help, but we focus on
the low-resource setting where only a small dictionary is available.

The Amharic labeled dataset is very small compared to other
languages because each Amharic example only contains one sentence.
As introduced in Section~\ref{ssec:objective}, one way to provide additional
training signal is by knowledge distillation from a third high-resource
language.
For the Amharic to Tigrinya \abr{cldc} experiment, we apply knowledge
distillation using English-Amharic parallel text.
We first train a reference English \abr{dan} on a large collection of labeled
English documents compiled from other \abr{lorelei} language packs.
We then use the knowledge distillation objective to train the \name{} models to
match the output of the English model on 1,200 English-Amharic parallel
documents sampled from the Amharic language pack.
To avoid introducing extra label bias, we sample the parallel documents such
that the English model output approximately follows a uniform distribution.

We do not use knowledge distillation on other language pairs.
For \abr{rcv2}, we already have enough labeled examples and therefore do not
need knowledge distillation.
For Tigrinya to Amharic \abr{cldc} experiment, we do not have enough unlabeled
parallel text in the Tigrinya language pack to apply knowledge distillation.

\subsection{Training Details}
\label{sec:hyperparameter}

For \abr{clwe}-based models, we use forty dimensional multiCCA
word embeddings~\citep{ammar-16}. 
We use three ReLU layers with 100 hidden units and 0.1 dropout for the
\abr{clwe}-based \abr{dan} models and the \abr{dan} classifier of the \name{}
models.
The \abr{bi-lstm} embedder uses ten dimensional character embeddings and forty
hidden states with no dropout.  The outputs of the embedder are forty
dimensional word embeddings.
We set $\lambda_d$ to 1, $\lambda_e$ to $0.001$, and $\lambda_p$ to 1 in the
multi-task objective~(Equation~\ref{eq:full_obj}).
The hyperparameters are tuned in a pilot Italian-Spanish \abr{cldc} experiment
using held-out datasets.

All models are trained with Adam~\citep{kingma-15} with default settings.
We run the optimizer for a hundred epochs with mini-batches of sixteen
documents.  For models that use additional resources, we also sample sixteen
examples from each type of training data (translation pairs, pre-trained
embeddings, or parallel text) to estimate the gradients of the auxiliary task
objectives $L_d$, $L_e$, and $L_p$ (defined in Section~\ref{ssec:objective}) at each
iteration.

\subsection{Effectiveness of \name{}}\label{ssec:analysis}
 
We train each model using ten different random seeds and report their average
test accuracy.
For models that use dictionaries, we also re-sample the training dictionary for
each run.
Table~\ref{tab:model} compares resource requirement and average \abr{rcv2}
accuracy of \name{} and baselines.
Table~\ref{tab:rcv2} and~\ref{tab:lorelei} show test accuracies on nine related
language pairs from $\abr{rcv2}$ and $\abr{lorelei}$.

\paragraph{Character-Level Knowledge Transfer.}
Experiments confirm that character-level knowledge transfer is sample-efficient
and complementary to word-level knowledge transfer.
The low-resource character-based \name{} models have similar average test
accuracy as the high-resource word-based models.
The \abr{src} variant does not use any
target language data, and yet its average test accuracy on \abr{rcv2} (50.0\%)
is very close to the \abr{clwe} model (51.6\%) and the supervised model
\abr{sup} (51.6\%).
When we already have a good \abr{clwe}, we can get the best of both worlds by
combining them (\abr{com}), which has a much higher average test accuracy
(64.5\%) than \name{} and the two baselines.

\paragraph{Multi-Task Learning.}
Training \name{} with multi-task learning further improves the accuracy.  For
almost all language pairs, the multi-task \name{} variants have higher test
accuracies than \abr{src}.
On \abr{rcv2}, word translation (\abr{dict}) is particularly effective even
with only 100 translation pairs. 
It increases average test accuracy from 50.0\% to 55.7\%, outperforming both
word-based baseline models.
Interestingly, word translation and mimick tasks together (\abr{all}) do not
consistently increase the accuracy over only using the dictionary (\abr{dict}).
On the \abr{lorelei} dataset where labeled document is limited, knowledge
distillation (\abr{src\pp{}} and \abr{mim\pp{}}) also increases
accuracies by around 1.5\%.

\paragraph{Language Relatedness.}
We expect character-level knowledge transfer to be less effective on language
pairs when the source language and the target language are less close to each
other.
For comparison, we experiment on \abr{rcv2} with transferring between more
distantly related language pairs: a North Germanic language and a Romance
language~(Table~\ref{tab:unrelated}).
Indeed, \name{} models score consistently
lower than the \abr{clwe}-based models when transferring from a North Germanic
source language to a Romance target language.  However, \name{} models are
surprisingly competitive with \abr{clwe}-based models when transferring from
the opposite direction.  This asymmetry is likely due to morphological
differences between the two language families.
Unfortunately, our datasets only have a limited number of language families.
We leave a more systematic study on how language proximity affect the
effectiveness of \name{} to future work.

\paragraph{Multi-Source Transfer.}
Languages can be similar along different dimensions, and therefore adding more
source languages may be beneficial.
On \abr{rcv2}, we experiment with training \name{} models on \emph{two} Romance
languages and testing on a third Romance language.
Moreover, using multiple source languages has a regularization
effect and prevents the model from overfitting to a single source language.
For fair comparison, we sample 750 training documents from each source
language, so that the multi-source models are still trained on 1,500 training
documents (like the single-source models).  We use a similar strategy to sample
the training dictionaries and pre-trained word embeddings.  Multi-source models
(Table~\ref{tab:multisrc}) consistently have higher accuracies than single-source models
(Table~\ref{tab:rcv2}).

\paragraph{Learned Word Representation.}
Word translation is a popular intrinsic evaluation task for cross-lingual 
word representations.
Therefore, we evaluate the word representations learned by the \abr{bi-lstm}
embedder on a word translation benchmark.
Specifically, we use the \abr{src} embedder to generate embeddings for all
French, Italian, and Spanish words that appear in multiCCA's vocabulary and 
translate each word with nearest-neighbor search.
Table~\ref{tab:bli} shows the top-1 word translation accuracy on the test
dictionaries from \abr{muse}~\citep{conneau-18}.
Although the \abr{src} embedder is not exposed to any cross-lingual signal, it
rivals \abr{clwe} on the word translation task by exploiting character-level
similarities between languages.

\paragraph{Qualitative Analysis.}
To understand how cross-lingual character-level similarity helps
classification, we manually compare the output of a \abr{clwe}-based model and
a \name{} model (\abr{dict} variant) from the Spanish to Italian \abr{cldc}
experiment.
Sometimes \name{} avoids the mistakes of \abr{clwe}-based models by correctly
aligning word pairs that are misaligned in the pre-trained \abr{clwe}.
For example, in the \abr{clwe}, ``relevancia'' (relevance) is the closest
Spanish word for the Italian word ``interesse'' (interest), while the
\abr{caco} embedder maps both the Italian word ``interesse'' (interest) and the
Spanish word ``interesse'' (interest) to the same point.  Consequently,
\abr{caco} correctly classifies an Italian document about the interest rate
with \abr{gcat} (government), while the \abr{clwe}-based model predicts
\abr{mcat} (market).


\section{Related Work}\label{sec:related}

Previous \abr{cldc} methods are typically word-based and rely on one of the
following cross-lingual signals to transfer knowledge: large bilingual
lexicons~\citep{shi-10,andrade-15}, \abr{mt} systems
~\citep{banea-08-fixed,wan-09-fixed,zhou-16}, or
\abr{clwe}~\citep{klementiev-12}.
One exception is the recently proposed multilingual
BERT~\citep{devlin-19,wu-19}, which uses a subword vocabulary.
Unfortunately, some languages do not have these resources.
\name{} can help bridge the resource gap.
By exploiting character-level similarities between related languages, \name{}
can work effectively with few or no target language data.

To adapt \abr{clwe} to low-resource settings, recent unsupervised \abr{clwe}
methods~\citep{conneau-18,artetxe-18b} do not use dictionary or parallel text.
These methods can be further improved with careful
normalization~\citep{zhang-19} and interactive refinement~\citep{yuan-19}.
However, unsupervised \abr{clwe} methods still require large monolingual
corpora in the target language, and they might fail when the monolingual corpora
of the two languages come from different domains~\citep{sogaard-18,fujinuma-19}
and when the two language have different morphology~\citep{czarnowska-19}.
In contrast, \name{} does not require any target language data.

Cross-lingual transfer at character-level is successfully used in low-resource
paradigm completion~\citep{kann-17}, morphological
tagging~\citep{cotterell-17a}, part-of-speech tagging~\citep{kim-17}, and named
entity recognition~\citep{bharadwaj-16,cotterell-17b,lin-18,rijhwani-19}, where
the authors train a character-level model jointly on a small labeled corpus in
target language and a large labeled corpus in source language.
Our method is similar in spirit, but we focus on \abr{cldc}, where it is less
obvious if orthographic features are helpful.
Moreover, we introduce a novel multi-task objective to use different types of
monolingual and cross-lingual resources.

\begin{table}
  \tabcolsep=0.15cm
  \centering
  \begin{tabular}{llcccc}
    \toprule
    source & target & \abr{src} & \abr{dict} & \abr{mim} & \abr{all} \\
    \midrule
    \flag{fr}~\flag{it}~\abr{fr}/\abr{it} & \flag{es}~\abr{es} & 58.8 & 67.0 & 55.8 & 65.3 \\
    \flag{es}~\flag{it}~\abr{es}/\abr{it} & \flag{fr}~\abr{fr} & 51.8 & 55.8 & 50.3 & 56.0 \\
    \flag{es}~\flag{fr}~\abr{es}/\abr{fr} & \flag{it}~\abr{it} & 53.2 & 56.1 & 55.9 & 56.5 \\
    \multicolumn{2}{r}{average} & 54.6 & 59.6 & 54.0 & 59.3 \\
    \bottomrule
  \end{tabular}
  \caption{Results of \abr{cldc} experiments using two source languages.
  Models trained on two source languages are generally better than models
  trained on only one source language (Table~\ref{tab:rcv2}).}
  \label{tab:multisrc} 
\end{table}

\begin{table}
  \centering
  \begin{tabular}{llRR}
    \toprule
    source & target & \abr{clwe} & \name{}\\
    \midrule
    \flag{es}~\abr{es} & \flag{fr}~\abr{fr} & 36.8 & 31.1\\
    \flag{es}~\abr{es} & \flag{it}~\abr{it} & 44.0 & 33.1\\
    \flag{fr}~\abr{fr} & \flag{es}~\abr{es} & 34.0 & 30.9\\
    \flag{fr}~\abr{fr} & \flag{it}~\abr{it} & 33.5 & 29.6\\
    \flag{it}~\abr{it} & \flag{es}~\abr{es} & 42.1 & 37.5\\
    \flag{it}~\abr{it} & \flag{fr}~\abr{fr} & 35.6 & 36.4\\
    \multicolumn{2}{r}{average} & 37.7 & 33.1\\
    \bottomrule
  \end{tabular}
  \caption{Word translation accuracies (P@1) for different embeddings.  The
  \name{} embeddings are generated by the embedder of a \abr{src} model trained
  on the source language.  Without any cross-lingual signal, the \name{}
  embedder has competitive word translation accuracy as \abr{clwe} pre-trained
  on large target language corpora and dictionaries.}
  \label{tab:bli} 
\end{table}

\section{Conclusion}\label{sec:conclusion}

We investigate character-level knowledge transfer between related languages for
\abr{cldc}.
Our transfer learning scheme, \name{}, exploits character-level similarities
between related languages through shared character representations to
generalize from source language data.
Empirical evaluation on multiple related language pairs confirm that
character-level knowledge transfer is highly effective.

\section*{Acknowledgement}
We thank the members of UMD CLIP and the anonymous reviewers for their
feedback.
Zhang and Boyd-Graber are supported by DARPA award HR0011-15-C-0113 under
subcontract to Raytheon BBN Technologies.
Fujinuma and Boyd-Graber are supported by NSF grant IIS-1564275.
Any opinions, findings, conclusions, or recommendations expressed here are
those of the authors and do not necessarily reflect the view of the sponsors.

\fontsize{9.0pt}{10.0pt}\selectfont
\bibliography{aaai}

\begin{thebibliography}{}

\bibitem[\protect\citeauthoryear{Ammar \bgroup et al\mbox.\egroup
  }{2016}]{ammar-16}
Ammar, W.; Mulcaire, G.; Tsvetkov, Y.; Lample, G.; Dyer, C.; and Smith, N.~A.
\newblock 2016.
\newblock Massively multilingual word embeddings.
\newblock {\em arXiv preprint arXiv:1602.01925}.

\bibitem[\protect\citeauthoryear{Andrade \bgroup et al\mbox.\egroup
  }{2015}]{andrade-15}
Andrade, D.; Sadamasa, K.; Tamura, A.; and Tsuchida, M.
\newblock 2015.
\newblock Cross-lingual text classification using topic-dependent word
  probabilities.
\newblock In {\em NAACL}.

\bibitem[\protect\citeauthoryear{Artetxe, Labaka, and
  Agirre}{2018}]{artetxe-18b}
Artetxe, M.; Labaka, G.; and Agirre, E.
\newblock 2018.
\newblock A robust self-learning method for fully unsupervised cross-lingual
  mappings of word embeddings.
\newblock In {\em ACL}.

\bibitem[\protect\citeauthoryear{Ballesteros, Dyer, and
  Smith}{2015}]{ballesteros-15}
Ballesteros, M.; Dyer, C.; and Smith, N.~A.
\newblock 2015.
\newblock Improved transition-based parsing by modeling characters instead of
  words with {LSTM}s.
\newblock In {\em EMNLP}.

\bibitem[\protect\citeauthoryear{Banea \bgroup et al\mbox.\egroup
  }{2008}]{banea-08-fixed}
Banea, C.; Mihalcea, R.; Wiebe, J.; and Hassan, S.
\newblock 2008.
\newblock Multilingual subjectivity analysis using machine translation.
\newblock In {\em EMNLP}.

\bibitem[\protect\citeauthoryear{Bharadwaj \bgroup et al\mbox.\egroup
  }{2016}]{bharadwaj-16}
Bharadwaj, A.; Mortensen, D.~R.; Dyer, C.; and Carbonell, J.~G.
\newblock 2016.
\newblock Phonologically aware neural model for named entity recognition in low
  resource transfer settings.
\newblock In {\em EMNLP}.

\bibitem[\protect\citeauthoryear{Chen \bgroup et al\mbox.\egroup
  }{2018}]{chen-18}
Chen, X.; Sun, Y.; Athiwaratkun, B.; Cardie, C.; and Weinberger, K.
\newblock 2018.
\newblock Adversarial deep averaging networks for cross-lingual sentiment
  classification.
\newblock {\em TACL} 6:557--570.

\bibitem[\protect\citeauthoryear{Collobert \bgroup et al\mbox.\egroup
  }{2011}]{collobert-11b}
Collobert, R.; Weston, J.; Bottou, L.; Karlen, M.; Kavukcuoglu, K.; and Kuksa,
  P.~P.
\newblock 2011.
\newblock Natural language processing (almost) from scratch.
\newblock {\em JMLR} 12:2493--2537.

\bibitem[\protect\citeauthoryear{Conneau \bgroup et al\mbox.\egroup
  }{2018}]{conneau-18}
Conneau, A.; Lample, G.; Ranzato, M.; Denoyer, L.; and J{\'e}gou, H.
\newblock 2018.
\newblock Word translation without parallel data.
\newblock In {\em ICLR}.

\bibitem[\protect\citeauthoryear{Cotterell and Duh}{2017}]{cotterell-17b}
Cotterell, R., and Duh, K.
\newblock 2017.
\newblock Low-resource named entity recognition with cross-lingual,
  character-level neural conditional random fields.
\newblock In {\em IJCNLP}.

\bibitem[\protect\citeauthoryear{Cotterell and Heigold}{2017}]{cotterell-17a}
Cotterell, R., and Heigold, G.
\newblock 2017.
\newblock Cross-lingual character-level neural morphological tagging.
\newblock In {\em EMNLP}.

\bibitem[\protect\citeauthoryear{Czarnowska \bgroup et al\mbox.\egroup
  }{2019}]{czarnowska-19}
Czarnowska, P.; Ruder, S.; Grave, E.; Cotterell, R.; and Copestake, A.
\newblock 2019.
\newblock Don't forget the long tail! a comprehensive analysis of morphological
  generalization in bilingual lexicon induction.
\newblock In {\em EMNLP}.

\bibitem[\protect\citeauthoryear{Devlin \bgroup et al\mbox.\egroup
  }{2019}]{devlin-19}
Devlin, J.; Chang, M.-W.; Lee, K.; and Toutanova, K.
\newblock 2019.
\newblock {BERT}: Pre-training of deep bidirectional transformers for language
  understanding.
\newblock In {\em NAACL}.

\bibitem[\protect\citeauthoryear{Fujinuma, Boyd-Graber, and
  Paul}{2019}]{fujinuma-19}
Fujinuma, Y.; Boyd-Graber, J.; and Paul, M.~J.
\newblock 2019.
\newblock A resource-free evaluation metric for cross-lingual word embeddings
  based on graph modularity.
\newblock In {\em ACL}.

\bibitem[\protect\citeauthoryear{Graves and Schmidhuber}{2005}]{graves-05}
Graves, A., and Schmidhuber, J.
\newblock 2005.
\newblock Framewise phoneme classification with bidirectional {LSTM} and other
  neural network architectures.
\newblock {\em Neural Networks} 18(5-6):602--610.

\bibitem[\protect\citeauthoryear{Iyyer \bgroup et al\mbox.\egroup
  }{2015}]{iyyer-15-fixed}
Iyyer, M.; Manjunatha, V.; Boyd-Graber, J.; and {Daum\'{e} III}, H.
\newblock 2015.
\newblock Deep unordered composition rivals syntactic methods for text
  classification.
\newblock In {\em ACL}.

\bibitem[\protect\citeauthoryear{Kann, Cotterell, and
  Sch{\"{u}}tze}{2017}]{kann-17}
Kann, K.; Cotterell, R.; and Sch{\"{u}}tze, H.
\newblock 2017.
\newblock One-shot neural cross-lingual transfer for paradigm completion.
\newblock In {\em ACL}.

\bibitem[\protect\citeauthoryear{Kim \bgroup et al\mbox.\egroup
  }{2017}]{kim-17}
Kim, J.-K.; Kim, Y.-B.; Sarikaya, R.; and Fosler-Lussier, E.
\newblock 2017.
\newblock Cross-lingual transfer learning for {POS} tagging without
  cross-lingual resources.
\newblock In {\em EMNLP}.

\bibitem[\protect\citeauthoryear{Kingma and Ba}{2015}]{kingma-15}
Kingma, D.~P., and Ba, J.
\newblock 2015.
\newblock Adam: A method for stochastic optimization.
\newblock In {\em ICLR}.

\bibitem[\protect\citeauthoryear{Klementiev, Titov, and
  Bhattarai}{2012}]{klementiev-12}
Klementiev, A.; Titov, I.; and Bhattarai, B.
\newblock 2012.
\newblock Inducing crosslingual distributed representations of words.
\newblock In {\em COLING}.

\bibitem[\protect\citeauthoryear{Lample \bgroup et al\mbox.\egroup
  }{2016}]{lample-16}
Lample, G.; Ballesteros, M.; Subramanian, S.; Kawakami, K.; and Dyer, C.
\newblock 2016.
\newblock Neural architectures for named entity recognition.
\newblock In {\em NAACL}.

\bibitem[\protect\citeauthoryear{Lewis \bgroup et al\mbox.\egroup
  }{2004}]{lewis-04}
Lewis, D.~D.; Yang, Y.; Rose, T.~G.; and Li, F.
\newblock 2004.
\newblock {RCV1}: A new benchmark collection for text categorization research.
\newblock {\em JMLR} 5(Apr):361--397.

\bibitem[\protect\citeauthoryear{Lin \bgroup et al\mbox.\egroup
  }{2018}]{lin-18}
Lin, Y.; Yang, S.; Stoyanov, V.; and Ji, H.
\newblock 2018.
\newblock A multi-lingual multi-task architecture for low-resource sequence
  labeling.
\newblock In {\em ACL}.

\bibitem[\protect\citeauthoryear{Ling \bgroup et al\mbox.\egroup
  }{2015}]{ling-15a}
Ling, W.; Dyer, C.; Black, A.~W.; Trancoso, I.; Fermandez, R.; Amir, S.;
  Marujo, L.; and Lu{\'{\i}}s, T.
\newblock 2015.
\newblock Finding function in form: Compositional character models for open
  vocabulary word representation.
\newblock In {\em EMNLP}.

\bibitem[\protect\citeauthoryear{Mikolov, Le, and
  Sutskever}{2013}]{mikolov-13b}
Mikolov, T.; Le, Q.~V.; and Sutskever, I.
\newblock 2013.
\newblock Exploiting similarities among languages for machine translation.
\newblock {\em arXiv preprint arXiv:1309.4168}.

\bibitem[\protect\citeauthoryear{Mimno \bgroup et al\mbox.\egroup
  }{2009}]{mimno-09-fixed}
Mimno, D.; Wallach, H.; Naradowsky, J.; Smith, D.; and McCallum, A.
\newblock 2009.
\newblock Polylingual topic models.
\newblock In {\em EMNLP}.

\bibitem[\protect\citeauthoryear{Mortensen, Dalmia, and
  Littell}{2018}]{mortensen-18}
Mortensen, D.~R.; Dalmia, S.; and Littell, P.
\newblock 2018.
\newblock Epitran: Precision {G2P} for many languages.
\newblock In {\em LREC}.

\bibitem[\protect\citeauthoryear{Pinter, Guthrie, and
  Eisenstein}{2017}]{pinter-17}
Pinter, Y.; Guthrie, R.; and Eisenstein, J.
\newblock 2017.
\newblock Mimicking word embeddings using subword {RNN}s.
\newblock In {\em EMNLP}.

\bibitem[\protect\citeauthoryear{Rijhwani \bgroup et al\mbox.\egroup
  }{2019}]{rijhwani-19}
Rijhwani, S.; Xie, J.; Neubig, G.; and Carbonell, J.~G.
\newblock 2019.
\newblock Zero-shot neural transfer for cross-lingual entity linking.
\newblock In {\em AAAI}.

\bibitem[\protect\citeauthoryear{Shi, Mihalcea, and Tian}{2010}]{shi-10}
Shi, L.; Mihalcea, R.; and Tian, M.
\newblock 2010.
\newblock Cross language text classification by model translation and
  semi-supervised learning.
\newblock In {\em EMNLP}.

\bibitem[\protect\citeauthoryear{S{\o}gaard, Ruder, and
  Vuli{\'c}}{2018}]{sogaard-18}
S{\o}gaard, A.; Ruder, S.; and Vuli{\'c}, I.
\newblock 2018.
\newblock On the limitations of unsupervised bilingual dictionary induction.
\newblock In {\em ACL}.

\bibitem[\protect\citeauthoryear{Strassel and Tracey}{2016}]{strassel-16}
Strassel, S., and Tracey, J.
\newblock 2016.
\newblock {LORELEI} language packs: Data, tools, and resources for technology
  development in low resource languages.
\newblock In {\em LREC}.

\bibitem[\protect\citeauthoryear{Wan}{2009}]{wan-09-fixed}
Wan, X.
\newblock 2009.
\newblock Co-training for cross-lingual sentiment classification.
\newblock In {\em ACL}.

\bibitem[\protect\citeauthoryear{Wu and Dredze}{2019}]{wu-19}
Wu, S., and Dredze, M.
\newblock 2019.
\newblock Beto, bentz, becas: The surprising cross-lingual effectiveness of
  {BERT}.
\newblock In {\em EMNLP}.

\bibitem[\protect\citeauthoryear{Xu and Yang}{2017}]{xu-17}
Xu, R., and Yang, Y.
\newblock 2017.
\newblock Cross-lingual distillation for text classification.
\newblock In {\em ACL}.

\bibitem[\protect\citeauthoryear{Yuan \bgroup et al\mbox.\egroup
  }{2019}]{yuan-19}
Yuan, M.; Zhang, M.; Durme, B.~V.; Findlater, L.; and Boyd-Graber, J.
\newblock 2019.
\newblock Interactive refinement of cross-lingual word embeddings.
\newblock {\em arXiv preprint arXiv:1911.03070}.

\bibitem[\protect\citeauthoryear{Yuan, Van~Durme, and
  Boyd-Graber}{2018}]{yuan-18}
Yuan, M.; Van~Durme, B.; and Boyd-Graber, J.
\newblock 2018.
\newblock Multilingual anchoring: Interactive topic modeling and alignment
  across languages.
\newblock In {\em NeurIPS}.

\bibitem[\protect\citeauthoryear{Zhang \bgroup et al\mbox.\egroup
  }{2019}]{zhang-19}
Zhang, M.; Xu, K.; Kawarabayashi, K.; Jegelka, S.; and Boyd-Graber, J.
\newblock 2019.
\newblock Are girls neko or sh\={o}jo? {C}ross-lingual alignment of
  non-isomorphic embeddings with {I}terative {N}ormalization.
\newblock In {\em ACL}.

\bibitem[\protect\citeauthoryear{Zhou, Wan, and Xiao}{2016}]{zhou-16}
Zhou, X.; Wan, X.; and Xiao, J.
\newblock 2016.
\newblock Cross-lingual sentiment classification with bilingual document
  representation learning.
\newblock In {\em ACL}.

\end{thebibliography}
\bibliographystyle{aaai}

\end{document}